\documentclass{article}

\usepackage{arxiv}

\usepackage[utf8]{inputenc} 
\usepackage[T1]{fontenc}    
\usepackage[pagebackref=true]{hyperref}       
\usepackage{url}            
\usepackage{booktabs}       
\usepackage{amsfonts}       
\usepackage{amsmath}
\usepackage{nicefrac}       
\usepackage{microtype}      
\usepackage{graphicx}
\usepackage{natbib}
\usepackage{lscape}
\usepackage{algpseudocode}
\usepackage{listings}
\usepackage{xargs}
\usepackage{float}
\usepackage{xcolor}
\usepackage{natbib}
\usepackage{doi}
\usepackage{xr-hyper}
\usepackage{algorithm2e}

\definecolor{highlight}{rgb}{0.90,0.88,0.88}
\definecolor{codegreen}{rgb}{0,0.6,0}
\definecolor{codeblue}{rgb}{0,0,0.6}
\definecolor{codegray}{rgb}{0.5,0.5,0.5}
\definecolor{codepurple}{rgb}{0.58,0,0.82}
\definecolor{backcolour}{rgb}{0.97,0.97,0.99}

\lstdefinestyle{mystyle}{
    backgroundcolor=\color{backcolour},
    commentstyle=\color{codegreen},
    keywordstyle=\color{codeblue},
    stringstyle=\color{codepurple},
    basicstyle=\ttfamily\small,
    breakatwhitespace=false,
    breaklines=true,
    captionpos=b,
    keepspaces=true,
    showspaces=false,
    showstringspaces=false,
    showtabs=false,
    tabsize=4,
	xleftmargin=0.5cm,frame=tlbr,framesep=4pt,framerule=0pt
}

\title{Pretraining the Vision Transformer using self-supervised methods for vision based Deep Reinforcement Learning}

\author{\href{https://orcid.org/0000-0001-6478-2038}{\includegraphics[scale=0.06]{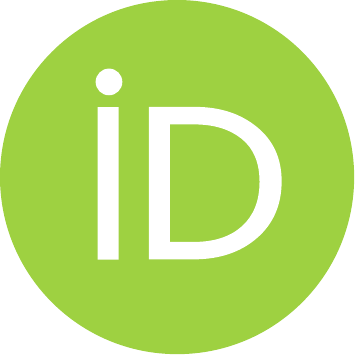}\hspace{1mm}Manuel~Goulão$^{a,b,c}$}
	\\
	\texttt{manuel.silva.goulao@tecnico.ulisboa.pt} \\
	\And \href{https://orcid.org/0000-0001-8638-5594}{\includegraphics[scale=0.06]{orcid.pdf}\hspace{1mm}
	Arlindo L.~Oliveira $^{a,b}$} \\
	\texttt{arlindo.oliveira@tecnico.ulisboa.pt}
}

\date{
    $^a$ Instituto Superior Técnico\\%
    $^b$ INESC-ID\\%
    $^c$ NeuralShift\\%
}

\renewcommand{\headeright}{Preprint}
\renewcommand{\undertitle}{Preprint}
\renewcommand{\shorttitle}{Pretraining the Vision Transformer using SSL for Deep RL}

\hypersetup{
pdftitle={Pretraining the Vision Transformer using self-supervised methods for vision based Deep Reinforcement Learning},
pdfsubject={},
pdfauthor={Manuel~Goulão},
pdfkeywords={},
}

\begin{document}
\maketitle

\begin{abstract}

The Vision Transformer architecture has shown to be competitive in the computer vision (CV) space where it has dethroned convolution-based networks in several benchmarks. Nevertheless, convolutional neural networks (CNN) remain the preferential architecture for the representation module in reinforcement learning. In this work, we study pretraining a Vision Transformer using several state-of-the-art self-supervised methods and assess the quality of the learned representations. To show the importance of the temporal dimension in this context we propose an extension of VICReg to better capture temporal relations between observations by adding a temporal order verification task. Our results show that all methods are effective in learning useful representations and avoiding representational collapse for observations from Atari Learning Environment (ALE) which leads to improvements in data efficiency when we evaluated in reinforcement learning (RL). Moreover, the encoder pretrained with the temporal order verification task shows the best results across all experiments, with richer representations, more focused attention maps and sparser representation vectors throughout the layers of the encoder, which shows the importance of exploring such similarity dimension. With this work, we hope to provide some insights into the representations learned by ViT during a self-supervised pretraining with observations from RL environments and which properties arise in the representations that lead to the best-performing agents. Source code will be available at: \url{https://github.com/mgoulao/TOV-VICReg}

\end{abstract}

\section{Introduction}

In recent years, a new architecture for vision-based tasks that does not use convolutions called the Vision Transformer (ViT) \citep{dosovitskiy_image_2020} has shown impressive results in several benchmarks. This architecture presents much weaker inductive biases when compared to a CNN, which can result in lower data efficiency. The Vision Transformer, unlike the CNNs, can capture relations between parts of an image (patches) that are far apart from each other, thus deriving global information that can help the model perform better in certain tasks. When the model is pretrained, using supervised or self-supervised learning, it manages to surpass in some cases the best convolution-based models in terms of task performance. Nonetheless, despite the successes obtained in computer vision these results are yet to be seen in reinforcement learning. Moreover, while some areas of machine learning have transitioned to large pretrained models, current Deep RL research is still largely based on small neural networks that are trained from \textit{tabula rasa}.

Despite the successes of deep reinforcement learning agents in the last decade, these still require a large amount of data or interactions to learn good policies. This data inefficiency makes current methods difficult to apply to environments where interactions are more expensive or data is scarce, which is the case in many real-world applications. In environments where the agent does not have full access to the current state, i.e. partially observable environments, this problem becomes even more prominent, since the agent not only needs to learn the state-to-action mapping but also a state representation function that tries to be informative about the current state given an observation. In contrast, humans, when learning a new task, already have a well-developed visual system and a good model of the world which are components that allow us to easily learn new tasks. Previous works have tried to tackle the sample inefficiency problem by using auxiliary learning tasks \citep{schwarzer_pretraining_2021, stooke_decoupling_2021, guo_bootstrap_2020}, that try to help the network's encoder to learn good representations of the observations given by the environments. These tasks can be supervised or unsupervised and can happen during a pretraining phase or during a reinforcement learning phase in a joint-learning or decoupled-learning scheme.

Recent results have shown that self-supervised learning is very useful in computer vision. Increased interest in this area has resulted in the appearance of new and improved methods that train a network to learn important features from the data using only the data itself as supervision. A common approach to evaluating such methods is to train a network composed of the pretrained encoder, with the parameters frozen, and a linear layer using popular datasets, like ImageNet. These evaluations have shown that these methods can achieve high scores in different benchmarks, which shows how well the current state-of-the-art methods are able to encode useful information from the given images without being task-specific. Additionally, it has been shown that pretraining a network using self-supervised learning (or unsupervised) adds robustness to the network and gives better generalization capabilities \citep{erhan_why_2010}.

Motivated by the potential of the Vision Transformer, in particular when paired with a pretraining phase, and the increasing interest in self-supervised tasks for DRL, we study pretraining ViT using state-of-the-art (SOTA) self-supervised learning methods for images. However, unlike images from datasets like ImageNet or MSCOCO, observations from reinforcement learning environments share similarities in more dimensions, for example, time \citep{stooke_decoupling_2021, anand_unsupervised_2020}, semantics \citep{fan2022minedojo,zhong2022improving}, and behavior \citep{agarwal_contrastive_2021}. To show the importance of these dimensions in comparison to current SOTA methods we propose extending VICReg (Variance Invariance Covariance Regularization) \citep{bardes_vicreg_2021} with a temporal order verification task \citep{misra_shuffle_2016} to help the model better capture the temporal relations between consecutive observations. We named this approach Temporal Order Verification-VICReg or in short TOV-VICReg. While we could have adapted any of the other methods, we opted for VICReg due to its computational performance, simplicity, and robustness against collapse.

We evaluate the different pretrained encoders in a data-efficiency regime and a linear probing task to determine which methods produce a better initialization for the model, assess if any pretrained model shows signs of representational collapse, and conduct a series of experiments to better understand the properties present in the representations. In our discussion, we also highlight some of the challenges that we faced during the experiments and propose some changes that can alleviate them.

Our main contributions are:
\begin{itemize}
	\item A proposal to combine two pretext tasks, VICReg and temporal order verification, to capture temporal relations between consecutive observations in reinforcement learning environments, in Section \ref{section:tov-vicreg}.
	\item The evaluation and comparison of the different self-supervised learning methods in Reinforcement Learning (Section \ref{section:data-efficiency}) and linear probing task (Section \ref{section:linear_probing}) based on imitation learning. The ViT pretrained with TOV-VICReg appears as the best performing model.
	\item A comparison of the different pretrained models using cosine similarity between the representations, attention maps and ratio of zeros in each layer of ViT (Section \ref{section:repr}). The results show that TOV-VICReg produces richer representations, more focused attention maps and sparser representation vectors.
\end{itemize}

\section{Related Work}

\paragraph{Vision Transformer for vision-based Deep RL} Recent work, has compared the Vision Transformer to convolution-based architectures with a similar number of parameters and shows that ViT is very data inefficient even when paired with an auxiliary task \citep{tao_evaluating_2022}.

\paragraph{Pretraining representations} Previous work has explored, similarly to our approach, pretraining representations using self-supervised methods which led to great data-efficiency improvements in the fine-tuning phase \citep{schwarzer_pretraining_2021,zhan_framework_2020} or superior results in evaluation tasks, like AtariARI \citep{anand_unsupervised_2020}. Others have pretrained representations using RL algorithms, like DQN, and transferred those learned representations to a new learning task \citep{wang_investigating_2022}.

\paragraph{Joint learning and augmentations} In recent years, adding an auxiliary loss to the RL loss, usually called joint learning, has become a common approach by many proposed methods. Curl \citep{srinivas_curl_2020} adds a contrastive loss using a siamese network with a momentum encoder. Another work studies different joint-learning frameworks using different self-supervised methods \citep{li_does_2022}. SPR \citep{schwarzer_data-efficient_2021} uses an auxiliary task that consists in training the encoder followed by an RNN to predict the encoder representation k steps into the future. PSEs \citep{agarwal_contrastive_2021} combines a policy similarity metric (PSM), that measures the similarity of states in terms of the behaviour of the policy in those states, and a contrastive task for the embeddings (CME) that helps to learn more robust representations. PBL \citep{guo_bootstrap_2020} learns representations through an interdependence between an encoder, which is trained to be informative about the history that led to that observation, and an RNN that is trained to predict the representations of future observations. Proto-RL \citep{yarats_reinforcement_2021} uses an auxiliary self-supervised objective to learn representations and prototypes \citep{caron_unsupervised_2020}, and uses the learned prototypes to compute intrinsic rewards that will push the agent to explore the environment.

A big contributor to the success of some joint learning methods has been the use of augmentations. Methods like DrQ \citep{kostrikov_image_2021} and RAD \citep{laskin_reinforcement_2020} pair an RL algorithm, like SAC, with image augmentations to improve data efficiency and generalization of the algorithms without using any auxiliary function.

\paragraph{Self-Supervised learning for image sequences}

Multiple works propose simple pretext tasks to train encoders to capture information from image sequences. These pretexts tasks can be playback speed classification \citep{yao2020video}, a temporal order classification \citep{misra_shuffle_2016, lee2017unsupervised, xu2019self}, a jigsaw game \citep{ahsan2019video} or a masked modelling task \citep{sun2019videobert}. A different approach consists of using contrastive learning. In this category, we can find methods that maximise the similarity between image sequences \citep{feichtenhofer2021large}, use autoregressive models to predict frames multiple steps in the future \cite{lorre2020temporal}, and maximize the similarity between temporally adjacent frames \citep{knights2021temporally}.

\section{Background}

\subsection{Vision Transformer}

ViT \citep{dosovitskiy_image_2020} is a model, for image classification tasks, that doesn't rely on CNNs and uses self-attention mechanisms. The model wraps the encoder of a Transformer by using linear projections of the patches extracted from the input image as tokens and adding a classification token which after the computation will serve as the image representation. When compared to CNNs, ViT presents weaker image-specific inductive biases which can impact the sample-efficiency of the model during learning \citep{dascoli_convit_2021}. However, it has been shown that with enough data the image-specific inductive biases become less important \citep{dosovitskiy_image_2020}. Moreover, ViT can capture relations between patches that are far apart from each other, thus deriving global information that can help the model perform better in certain tasks.

\subsection{Reinforcement Learning}

The problem of an \textbf{agent} learning to solve a task in a certain \textbf{environment} can be defined as a Markov Decision Process (MDP). A MDP $\mathcal{M}$ is defined by the tuple $\left \langle \mathcal{S},\mathcal{A},\mathcal{R},\mathcal{T}  \right \rangle$, where $\mathcal{S}$ is the set of states, $\mathcal{A}$ the set of actions, $\mathcal{R}$ the reward function, and $\mathcal{T}$ the transition function. At each timestep the agent is in a state $s\in \mathcal{S}$ and takes an action $a \in \mathcal{A}$. Upon performing the action the agent receives from the environment a reward $r\in\mathcal{R}$ and a new state $s^\prime\in \mathcal{S}$ which is determined by the transition function $\mathcal{T}(s^\prime, s, a)$. The MDP assumes that the Markov property holds in the environment, i.e. that the state transitions are independent and the agent only needs to know the current state to perform an action $P(a_t|x_0,x_1 ...x_t)=P(a_t|x_t)$. For the agent to decide what action to take it uses a policy function $\pi$, which gives a distribution over actions given a state, $\pi(a_t|s_t)$. This policy is evaluated using the function $V^\pi(s)$, which estimates the expected total discounted reward of an agent in a state $s$ and that follows a policy $\pi$.

\subsubsection{DQN and Rainbow}

DQN \citep{mnih_playing_2013} is a value-based method and uses a network with parameters $\phi$ that given a state $s$ outputs a prediction of the distribution of Q values over actions, $Q_\phi(s,a)$. The network learns the Q function by minimizing the mean squared error: $(y-Q_\phi(s,a))^2$, where $y=r +\gamma \ max_{a^\prime} Q_\phi(s^\prime,a^\prime)$, as shown in Algorithm \ref{alg:dqn} at the Appendix.

Several works followed the DQN algorithm which introduced changes to improve performance. Rainbow \citep{hessel_rainbow_2017} combines six improvements, Double Q-Learning \citep{van_hasselt_deep_2016}, Prioritized Replay \citep{schaul_prioritized_2016}, Dueling Networks \citep{wang_dueling_2016}, Multi-step Learning \citep{sutton_reinforcement_2018}, Distributional RL \citep{bellemare_distributional_2017}, and Noisy Nets \citep{fortunato_noisy_2018} resulting in a more stable and sample efficient algorithm.

\subsection{Self-Supervised methods}

For this study we selected DINO \citep{caron_emerging_2021}, MoCo \citep{chen_empirical_2021}, MAE \citep{he_masked_2021}, and VICReg \citep{bardes_vicreg_2021} since they are currently considered state-of-the-art, their official implementations are available in PyTorch, and each represents a different type of approach. MoCo \citep{he_momentum_2020} is a contrastive learning method meaning that it learns using a loss function that pulls the positive samples together and pushes the negative samples apart. MoCo, in particular, has three versions, 1 \citep{he_momentum_2020}, 2 \citep{chen_improved_2020}, and 3\citep{chen_empirical_2021}. In this work, we consider the most recent version (v3). On the other hand, non-contrastive methods (also called regularized) don't rely on the notion of positive and negative samples and only attempt to push different views from the same source together. To avoid collapse these methods use a set of tools that act as regularization, e.g. stop gradient, strong augmentations, and asymmetric siamese networks. From this class of methods, we consider DINO \citep{caron_emerging_2021} and VICReg. Lastly, we also consider MAE \citep{he_masked_2021}, a masked reconstruction method, which consists in training an auto-encoder based on ViT to reconstruct an image with a set of patches masked.

\section{TOV-VICReg}
\label{section:tov-vicreg}

VICReg is a non-contrastive method that trains a network to be invariant to augmentations applied to the inputs while avoiding a trivial solution with the help of two additional losses, called variance and covariance, that act as regularizers over the embeddings. While VICReg is agnostic concerning the architectures used and even the weight sharing, in this work we consider the version where paths are symmetric, the weights are shared, and each path is composed of an encoder (also called backbone) and an expander. The expander is a network that increases the dimension of the representation vector in a non-linear way allowing the covariance loss to reduce dependencies and not only correlations of the representation vector. In addition, the expander also removes information that is not common to both representations.

VICReg uses three loss functions: \textbf{invariance} is the mean of the square distance between each pair of embeddings from the same original image, as shown in Equation \ref{eqn:invariance}, where $Z$, and $Z^\prime$ are two sets of embeddings, of size $N$, that result from computing two different augmentations of $N$ sources, and $z_j$ denotes the \textit{j-th} embedding in the set; \textbf{variance} is a hinge loss that computes, over the batch, the standard deviation of the variables in the embedding vector and pushes that value to be above a certain threshold, as shown in Equation \ref{eqn:variance}, where $d$ denotes the number of dimensions of the embedding vector, and $Z^j$ is the set of the \textit{j-th} variables in the set of embedding $Z$; \textbf{covariance} is a function that computes the sum of the squared off-diagonal coefficients of a covariance matrix computed over a batch of embeddings, as shown in Equation \ref{eqn:covariance}, to decorrelate the variables from the embedding. While the invariance loss function tries to make the model invariant to augmentations, i.e. output the same representation vector, the other two functions act as regularizers by pushing the variables of the embedding vector to vary above a certain threshold and decorrelating the variables in each embedding vector.

\begin{equation}
	\label{eqn:invariance}
	i(Z,Z^\prime)=\frac{1}{N}\sum_{j}^{N}\left \|z_j - z^\prime_j \right \|_2^2
\end{equation}

\begin{equation}
	\label{eqn:variance}
	v(Z)=\frac{1}{d}\sum^{d}_{j}\textup{max}(0,\gamma-\sqrt{Var(Z^j)})
\end{equation}

\begin{equation}
	\label{eqn:covariance}
	c(Z)=\frac{1}{d}\sum_{i\neq j}\left [  \textup{Cov(Z)}\right ]^2_{i,j}
\end{equation}

\begin{figure}[h]
	\centering
    \includegraphics[width=9cm]{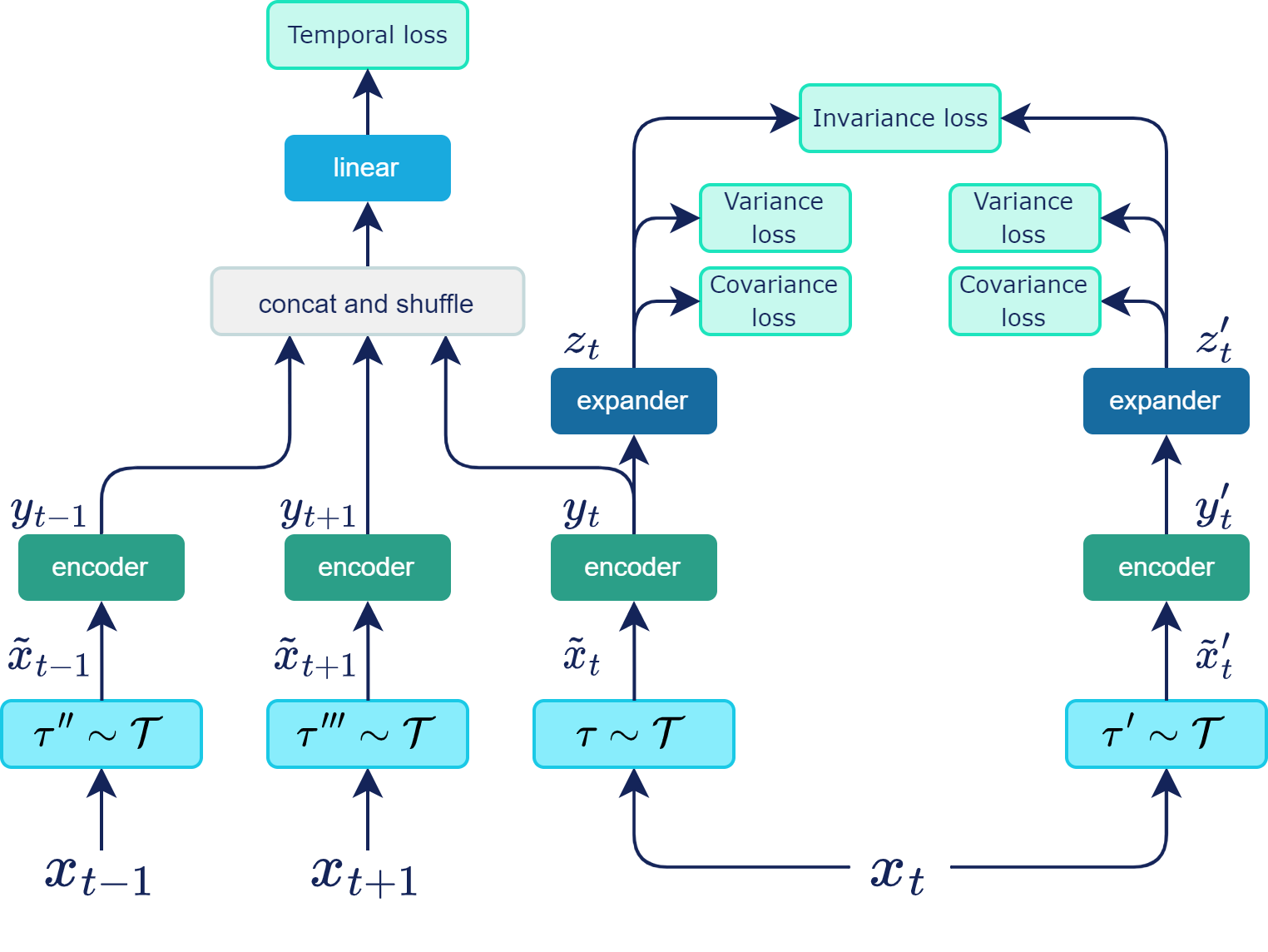}
	\caption{TOV-VICReg architecture}
	\label{fig:tov-vicreg}
\end{figure}

TOV-VICReg or Temporal-Order-Verification-VICReg extends VICReg to better capture the temporal relations between consecutive observations and consequently encode extra information that can be useful in the deep reinforcement learning phase. To achieve that we add a new temporal order verification task, as proposed at Shuffle-and-Learn \citep{misra_shuffle_2016}, that consists of a binary classification task where a linear layer learns to predict if three given representation vectors are in the correct order or not. Like the other losses, we also employ a coefficient for the temporal loss and in most of our experiments, the value is 0.1. Figure \ref{fig:tov-vicreg} visually illustrates TOV-VICReg.

At each step we sample three consecutive observations, $\{x_{t-1},\ x_t,\ x_{t+1}\}$. $x_t$ is processed by two different augmentations, and like VICReg these are the augmentations used in BYOL \citep{grill_bootstrap_2020}. $x_{t-1}$ and $x_{t+1}$ are processed by two simple augmentations composed of a color jitter and a random grayscale. The $x_t$ augmentations are computed by the VICReg computation path and the resultant embeddings are used with VICReg loss functions, i.e. variance, invariance, and covariance. For the \textbf{temporal order verification task} we encode the augmentation of $x_{t-1}$ and $x_{t+1}$, and concatenate those two representations with one of the representations of $x_t$. In our case we used the one that was augmented without solarize, obtaining the vector $\{y_{t-1},y_t,y_{t+1}\}$.  Finally, we randomly permute the order of the representations in the vector and feed the resultant concatenated vector to a linear layer with a single output node that predicts if the given concatenated vector has the representations in the right order or not. The \textbf{temporal loss} used to optimize the model for this task is a Binary Cross Entropy loss. TOV-VICReg's pseudocode can be found in Appendix \ref{sec:tov-vicreg-pseudo}.

\section{Pre-Training Methodology}

We pretrained five encoders, one using our proposed method TOV-VICReg and four using state-of-the-art self-supervised methods: MoCo v3 \citep{chen_empirical_2021}, DINO \citep{caron_emerging_2021}, VICReg \citep{bardes_vicreg_2021}, and MAE \citep{he_masked_2021}.
For this study, the encoder used is a Vision Transformer, more precisely the ViT tiny. We use a patch size of 8 for all SSL methods except MAE where the value is 7, since it requires the observation size (84) to be divisible by the patch size. In Appendix \ref{sec:vit_patch} we further explore the choice of these values for the patch size and show that the patch size of 7 used for MAE does not affect the results. Moreover, the implementation we use is an adaptation of the timm library \citep{rw2019timm} implementation, which can be found in the source code of the DINO method. The dataset used is a set of observations from 10 of the 26 games in the Atari 100k benchmark, all available in the DQN Replay Dataset \citep{agarwal_optimistic_2020}. For each game, we use three checkpoints (1,25,50) with a size of one hundred thousand data points (observations), which makes up a total of three million data points (\textasciitilde 55 hours). The pretraining phase is 10 epochs with two warmup epochs. We used the official code bases of all the self-supervised methods and tried to change the least amount of hyperparameters. Appendix \ref{section:hyper} contains the tables with the hyperparameters used for each method.

\section{Representations Evaluation}

To evaluate the pretrained Vision Transformers we perform two experiments. In the first experiment, we evaluate the pretrained representations in a reinforcement learning setting and compare the data-efficiency gains. In the second experiment, we evaluate the pretrained representations using a linear probing task based on imitation learning.

\subsection{Data-Efficiency in RL}
\label{section:data-efficiency}

To evaluate the pretrained Vision Transformers in reinforcement learning and compare data-efficiency gains, we trained in the 10 games used for pre-training for 100k steps using the Rainbow algorithm \citep{hessel_rainbow_2017}, with the DER \citep{van_hasselt_when_2019} hyperparameters. The only difference between the agents at the start is the representation module. We chose two networks to compare against, the Nature CNN \citep{mnih_human-level_2015}, and a ResNet that has a similar number of parameters to ViT tiny (Appendix \ref{sec:resnet}) that has a number of parameters similar to the ViT tiny. Moreover, we use a learning rate two orders of magnitude smaller for the encoder ($1 \times 10^{-6}$), which previous works \citep{schwarzer_pretraining_2021} and experiments performed by us have shown to be beneficial. To report our results we follow the rliable \citep{agarwal_deep_2021} evaluation framework, where the scores of all games are normalized and treated as one single task.

\begin{figure}[H]
	\centering
    \includegraphics[width=14cm]{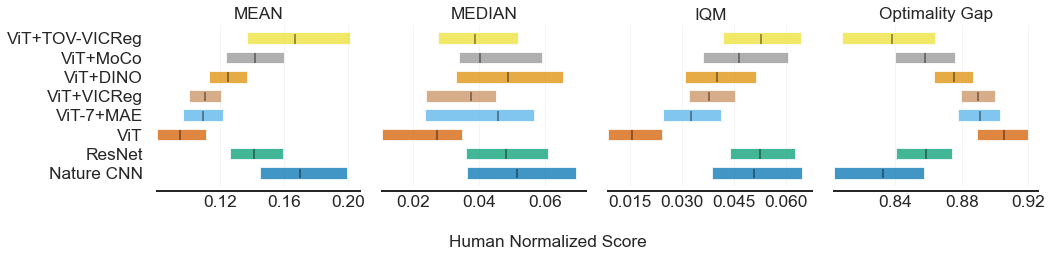}
	\caption{The eval runs across the different games are normalized and treated as a single task. The IQM corresponds to the Inter-Quartile Mean among all the runs, where the top and bottom 25\% are discarded and the mean is calculated over the remaining 50\%. The Optimality Gap refers to the number of runs that fail to surpass the human average score, i.e. 1.0.  }
	\label{fig:results}
\end{figure}

Figure \ref{fig:results} shows the aggregate metrics of seven different encoders on 10 Atari games with training runs of 100k steps. The first five (ViT+<method>) are ViT tiny models pretrained with five different self-supervised methods, while the last three (ViT, ResNet, and Nature CNN) are randomly initialized models.

Starting with the randomly initialized models we can assess that the Nature CNN and the ResNet are the most sample efficient models, with ViT far behind.

Regarding the pretrained encoders, ViT, when pretrained with TOV-VICReg, performs better than the other pretrained encoders and the non-pretrained ViT in all metrics except in the median.

The observed difference between the behavior of the mean and the median is explained by the fact that the distribution of scores obtained by TOV-VICReg has a long tail to the right. In fact, TOV-VICReg is the method that more commonly exhibits behavior that surpasses human performance, as shown by the optimality gap, pointing to the possibility that in some fraction of the cases, it finds good representations that allow the agent to learn a good policy faster. It is worth noting that we found a higher variance in the results of our proposed method, when compared to the remaining methods and non-pretrained models.

All self-supervised methods prove to be effective in improving the data-efficiency of ViT with MoCo showing the best results in IQM among the SOTA methods, followed by DINO, VICReg and MAE, respectively.

ViT+TOV-VICReg when compared to Nature CNN, which has far fewer parameters, and ResNet, with a similar number of parameters, seems to closely match their sample-efficiency performance (Appendix Table \ref{tab:model_size}). Furthermore, the difference between the ViT+TOV-VICReg and ViT+VICReg shows that exploring temporal relations results in better representations. Lastly, comparing the ViT+TOV-VICReg with the non-pretrained ViT shows that a good self-supervised method with 3 million data points can help close the sample-efficiency gap while remaining a more complex and capable model.

\subsection{Linear Probing}
\label{section:linear_probing}

Evaluating representations computed by a pretrained encoder is a difficult task. One possible option is assessing improvements in data efficiency in a reinforcement learning task, as we did in the previous section. However, the results usually suffer from a high level of uncertainty which requires us to run dozens of training runs, thus making it computationally expensive. Another possible path would be using previously proposed benchmarks like the AtariARI benchmark \citep{anand_unsupervised_2020}, which tries to evaluate representations using the RAM states as ground truth labels. However, this only works for 22 Atari games (out of 62) and requires the encoder to use the full observation provided by the environments (160x210). For those reasons, we use a different evaluation task that is more efficient, allowing us to test more pretrained models during the research process ($\sim 50$ min per game), and flexible, meaning that we can use it in different environments.

Our second experiment consists of linear probing the pretrained encoders in an imitation learning task. We present the results in Table \ref{tab:eval}, where we compare the pretrained encoders against a random classifier, i.e. uniform sampling, a randomly initialized ViT and a non-frozen encoder which we use as a goal. The results are aligned with the results from the previous section with all methods showing improvements in comparison to the randomly initialized ViT and once again ViT+TOV-VICReg showing better performance than the remaining pretrained encoders. Additional information about this evaluation task can be found in Appendix \ref{section:eval_complete}.

\begin{table}[h]
    \centering
    \scriptsize
    \begin{tabular}{p{0.12\textwidth}p{0.059\textwidth}|p{0.06\textwidth}|p{0.06\textwidth}p{0.06\textwidth}p{0.06\textwidth}p{0.06\textwidth}p{0.09\textwidth}|p{0.06\textwidth}}
        \toprule
        & & Random init & \multicolumn{5}{c|}{Pre-trained encoders} & \multicolumn{1}{c}{W/o freeze} \\
        \midrule
		& 				 Random Classifier & 	 ViT 	&  ViT+\newline DINO & ViT+\newline MoCo & ViT+\newline VICReg & ViT-7+\newline MAE  &	ViT+\newline TOV-VICReg & Nature CNN	 \\
		Game           &                   &            &        		 &          &          &            &          		   &                 \\
		\midrule
		Alien          &            0.0556 &     0.0147 &        0.0470 &   0.0646 &     0.0695 &           0.0988 &  0.1003 &   \textbf{0.1021} \\
		Assault        &            0.1519 &     0.1770 &         0.2536 &   0.2557 &     0.3704 &           0.3065 &  0.3044 &  \textbf{0.6673} \\
		BankHeist      &            0.0608 &     0.0756 &         0.1059 &   0.1083 &     0.1467 &           0.1523 & 0.1622 &   \textbf{0.2080} \\
		Breakout       &            0.2509 &     0.2183 &         0.3591 &   0.2765 &     0.4077 &           0.3099 & 0.3285 &   \textbf{0.5907} \\
		Chopper Command&            0.0563 &     0.0176 & 0.0383 &   0.2019 &     0.1298 &   		0.3088 & \textbf{0.3225}&   0.2660 \\
		Freeway        &            0.3999 &     0.6843 &         0.6850 &   0.6972 &     0.6971 &           0.6942 & 0.7041 &   \textbf{0.8885} \\
		Frostbite      &            0.0565 &     0.0367 & 0.0517 &  	0.0744 &     0.0664 &  			0.1001 & \textbf{0.1021}&   0.1019 \\
		Kangaroo       &            0.0603 &     0.0562 &        0.0877 &   0.1374 &     0.1259 &           0.0737 &  0.2184 &   \textbf{0.3311} \\
		MsPacman       &            0.1121 &     0.0780 &        0.1215 &   0.1168 &     0.1400 &           0.1500 &  0.1527 &   \textbf{0.2063} \\
		Pong           &            0.1644 &     0.0718 &         0.1447 &   0.2730 &     0.2337 &           0.1223 & 0.2853 &   \textbf{0.4340} \\
		\midrule
		Mean           &            0.1369 &     0.1430 &        0.1894 &   0.2206 &     0.2387 &           0.1706 &  0.2680 &   \textbf{0.3796} \\
		\bottomrule
    \end{tabular}
    \caption[Evaluation task results]{F1-scores for each game evaluated and mean of the F1-scores. We trained all the encoders in all games separately for 100 epochs over a dataset of 100k observations and evaluate 10k unseen observations. The rightmost column shows the results of a Nature CNN encoder that was not frozen during the training phase and which we use as a goal for the remaining.}
    \label{tab:eval}
\end{table}

\section{Collapse Evaluation}
\label{section:metrics}

A significant phenomenon when doing self-supervised training is the collapse of the representations, which can be seen in three forms: representational collapse, dimensional collapse, and informational collapse. Representational collapse refers to the features of the representation vector collapsing to a single value for every input, leading to a variance of the features of zero, or close to zero. In dimensional collapse, the representations don't use the full representation space, which can be measured by calculating the singular values of the covariance matrix calculated over the representations. Informational collapse corresponds to the case where the features of the representation vector are correlated and therefore are representing the same information.

\paragraph{Dimensional Collapse}

\begin{figure}[h]
	\centering
    \includegraphics[width=6cm]{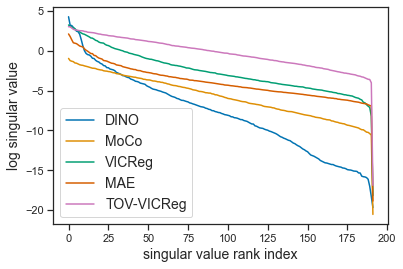}
	\caption{Logarithm of the singular values of the representation vector's covariance matrix sorted by value.}
	\label{fig:dimensional}
\end{figure}

All methods seem to avoid dimensional collapse, since most dimensions have a singular value larger than zero, as observed in Figure \ref{fig:dimensional}. However, we notice that some methods make better use of the space available since they present higher singular values. TOV-VICReg, in particular, seems to excel in this metric, even improving the results obtained by VICReg. It is worth noting that both VICReg and TOV-VICReg employ a covariance loss that helps decorrelate the embedding variables which may be contributing positively to these results. Furthermore, we used a covariance coefficient of 10 for TOV-VICReg and 1 for VICReg a change that according to our experiments culminates in the increase here observed.

\paragraph{Representational Collapse}

Results in the first row of Table \ref{tab:repr_collapse} show the computed standard deviation of the representation vector over a batch of thousands of data points. DINO, VICReg and TOV-VICReg show a value well above zero, meaning that none of the methods suffered from representation collapse during training. On the other hand, MoCo shows a much smaller value of 0.178, which is still, is far from a complete collapse. Both VICReg and TOV-VICReg use a hinge loss that pushes the representation vector to have a standard deviation of 1 or above. While VICReg slowly converges to this value our method converges to roughly 1.65, which might be the result of adding a temporal order verification task.

\begin{table}[ht]
	\centering
	\begin{tabular}{llllll}
		\toprule
		Metric & DINO  & MoCo  & VICReg & MAE & TOV-VICReg \\
		\midrule
		Std & 0.979 & 0.178 & 1.003 & 0.475 & 1.648 \\
		Corr. Coef. & 0.1764 & 0.1538 & 0.1531 & 0.1602 & 0.0780 \\
		\bottomrule
	\end{tabular}
	\caption{Average standard deviation and correlation coefficient of the representation vector}
	\label{tab:repr_collapse}
\end{table}

\paragraph{Informational Collapse}

We report in the second row of Table \ref{tab:repr_collapse}, the comparison of the average correlation coefficients of the representation vectors. TOV-VICReg performs better than the other methods, including VICReg, all of them with very similar coefficients. Like in the dimensional collapse, this result is in part due to the higher covariance coefficient used in TOV-VICReg which by design helps the model to decorrelate the representation's features. Increasing the coefficient in VICReg results in a lower correlation coefficient as well, but is still higher than TOV-VICReg.

\section{Analysis of the Representations}
\label{section:repr}

In this section, we present different visualizations to better understand the representations learned by each of the pretrained encoders. Our goal with the following visualizations is to help us better understand the learned representations, give some intuitions about their properties, and understand which properties are present in the encoders that performed better in Section \ref{section:data-efficiency} and \ref{section:linear_probing}.

\paragraph{Cosine similarity}

\begin{figure}[h]
	\centering
    \includegraphics[width=13cm]{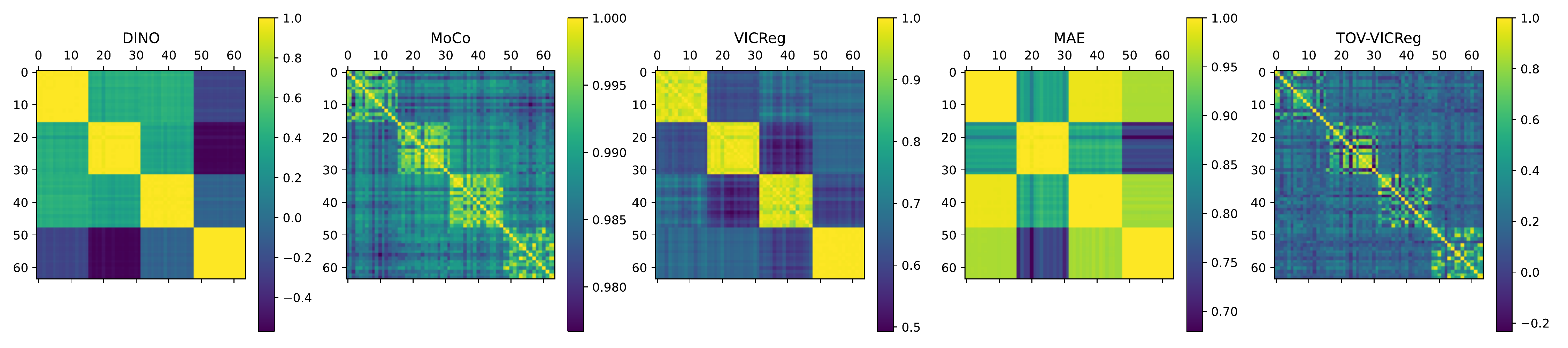}
	\caption{Similarity matrices of the representations computed by MoCo, DINO, VICReg, MAE and TOV-VICReg respectively. There are a total of 64 data points, from 4 different games: Alien, Breakout, MsPacman, and Pong, where from 0-15 are from Alien, 16-31 are from Breakout and so forth. }
	\label{fig:cosine-all}
\end{figure}

\begin{figure}[h]
	\centering
    \includegraphics[width=13cm]{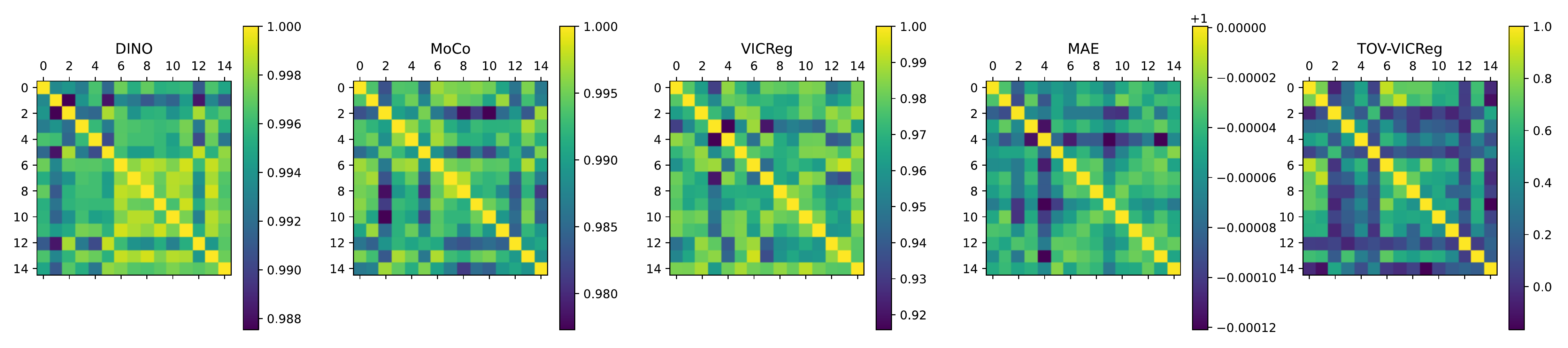}
	\caption{Similarity matrices of the representations computed by MoCo, DINO, VICReg, MAE and TOV-VICReg respectively, of observations from MsPacman. }
	\label{fig:cosine-game}
\end{figure}

Figure \ref{fig:cosine-all} presents a similarity matrix of the representations where we can observe that TOV-VICReg can better distinguish between observations of different games but also observations from the same game, as shown in Figure \ref{fig:cosine-game}. MoCo, on the other hand, seems to make a good distinction between observations from the different games. However, as we can observe in the colour bar, all the representations are very similar to each other, which corroborates the results obtained in Section \ref{section:metrics}. Oppositely, VICReg and DINO manage to spread representations more, as we can see in the colour bars, but, the yellow squares in the diagonal show that the representations from the same game are more similar to each other which is corroborated by Figure \ref{fig:cosine-game}. Given the empirical results, we believe that this capacity to distinguish observations from the same game might be a good indicator.

\paragraph{Attention visualisation}

\begin{figure}[h]
	\centering
    \includegraphics[width=8cm]{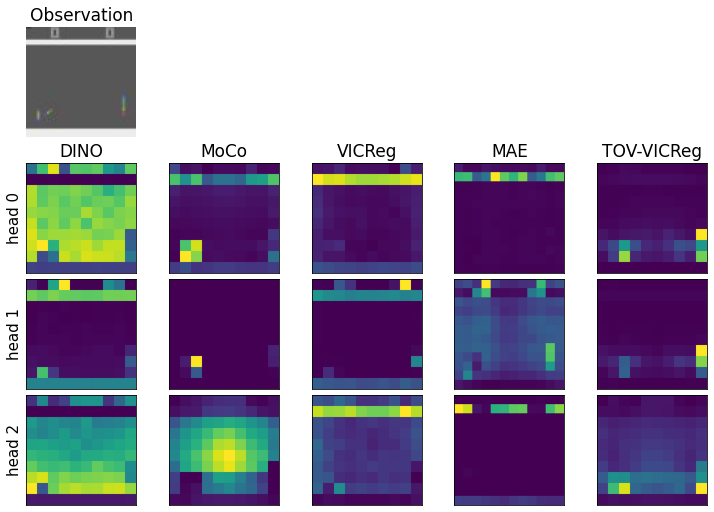}
	\caption{Attention maps produced by the pretrained ViTs. We fed a pretrained ViT with an observation from the game Pong and obtained the attention maps from the three heads in the last block. }
	\label{fig:attention}
\end{figure}

Inspired by the results presented in the DINO work \citep{caron_emerging_2021}, we perform an analysis of the attention maps of the different pretrained encoders. In Figure \ref{fig:attention}, we can see the results of all methods for an observation from the game of Pong, where each method produces three attention maps, one for each self-attention head of the last block of the Vision Transformer. All pretrained ViT seem to attend at some level to important game features like the ball and the paddles. However, TOV-VICReg is the only method that doesn't spread the attention to other parts of the frame that we don't consider important to describe the current state of the game. When comparing to VICReg's attention maps we believe that the temporal order verification task greatly helped the attention of the model. In more visually complex games, e.g. Freeway or MsPacman, these attention maps start to be more difficult to analyze but it is still possible to discern some important features.

\paragraph{Sparsity}

\begin{figure}[h]
	\centering
    \includegraphics[width=8cm]{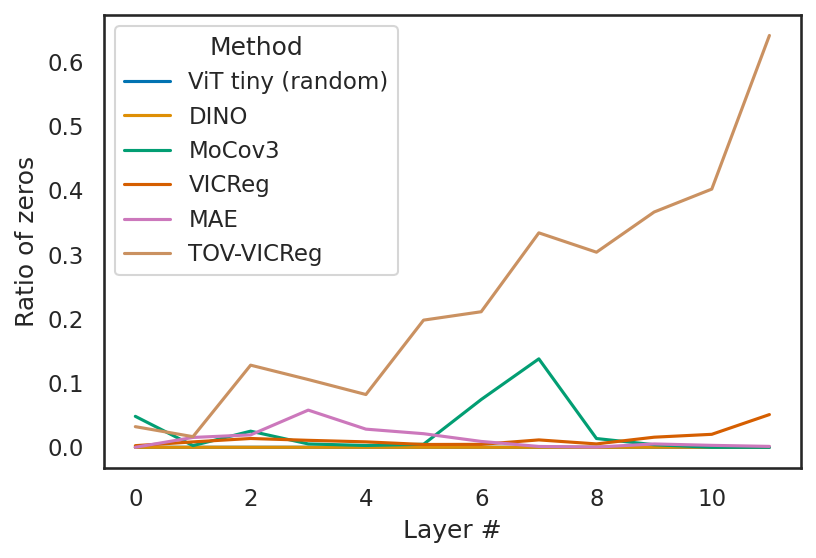}
	\caption{The ratio of zeros of the representation vector after the activation of the MLP across the different layers of Vision Transformer. }
	\label{fig:sparsity}
\end{figure}

Figure \ref{fig:sparsity} shows the ratio of zeros of the representation vector after the activation of the MLP across the different layers of Vision Transformer. We can see that TOV-VICReg has a higher sparsity than the other methods and that the sparsity increases after each layer of the network. Sparsity has been exploited to scale transformers to larger sizes while maintaining a reasonable number of floating point operations.

\section{Discussion \& Conclusion}

In this work, we presented a study of ViT for vision-based deep reinforcement learning using self-supervised pretraining, and proposed a simple self-supervised learning method that extends VICReg to better capture temporal relations between consecutive observations.  This type of approach has seen successes in natural language processing \citep{devlin_bert_2019, brown_language_2020}, and computer vision \citep{radford_learning_2021} and we believe that similar approaches in RL have the potential to unlock new levels of performance never achieved before \citep{baker_video_2022}. With this work, we hope to contribute to the growing body of work on self-supervised learning for RL and to provide important insights to the community on the importance of exploring dimensions where observations are similar and a better understanding of the representations learned during the self-supervised pretraining.

Our results showed that pretraining a Vision Transformer using SOTA self-supervised learning methods is effective in improving the data-efficiency of RL agents and improving the performance of a linear probe of the encoder in an imitation learning task. Additionally, we showed that exploring the temporal relations between consecutive observations can further improve the results, as the encoder pretrained with TOV-VICReg was the best performing in both experiments. We used several metrics to assess if the pretrained encoders suffered from representational collapse and found that all methods were effective in avoiding this problem and that TOV-VICReg shows the best results, especially in the use of the representation vector dimensions and low level of correlation between variables of the representation vector. Moreover, our analysis of the representations shows that the best-performing encoders were also the ones with richer representations in the cosine similarity matrix, more focused attention maps and a higher sparsity.
Sparsity in particular is a property that has been exploited to achieve better inference time and memory usage in the deployment of large Transformers \citep{jaszczur2021sparse, gupta2021memoryefficient} and which can be important to successfully deploy reinforcement learning agents in real-world applications that use much more capable models.

Considering the impact of the temporal dimension on our results we believe that future works may improve these results by exploring other dimensions, like, semantics and behavior. Another type of evaluation where this kind of approach has the potential to be extremely important and which we do not explore in this work is the generalization to unseen tasks, which can be different in different dimensions like observation and state.

Using models such as the Vision Transformer which are predominantly larger than the models commonly used in RL and therefore slower to train was a great challenge during this work and we believe it is necessary to push for a change in some of the common practices in the field of reinforcement learning. Firstly, when training agents online it is necessary to use paralyzed environments like EnvPool \citep{weng2022envpool} instead of running in a single processing environment. Additionally, as the SOTA progresses to larger and more robust models we will need environments that are capable to evaluate such capabilities. While ALE is still a valid option we believe that environments specifically designed for RL, like Procgen \citep{cobbe_leveraging_2020}, are preferable and are, in our opinion, still missing. In the context of pretrained models for reinforcement learning, we have found the linear probing evaluation task an extremely valuable approach to evaluate the quality of the pretrained models in a fast and informative way. For those reasons, we find the adoption of such evaluations and the development of new ones of great value for advancing the field.

Lastly, while our best pretrained encoder was only able to match the sample efficiency of a Nature CNN we were able to achieve a good improvement in comparison to the non-pretrained Vision Transformer. The ability to use larger models, with millions of parameters, that are as sample efficient as some of the most popular CNN-based models (like Nature CNN or Impala ResNet), with thousands of parameters, can open the door to using Deep RL in even more complex problems where smaller models tend to struggle to perform, without losing sample-efficiency.

\section{Acknowledgments}

We acknowledge the financial support provided by Recovery and Resilience Fund towards the Center for Responsible AI project (Ref. C628696807-00454142), the Foundation for Science and Technology (FCT) through the Project PRELUNA - PTDC/CCI-INF/4703/2021  and also the multiannual financing of the Foundation for Science and Technology (FCT) for INESC-ID (Ref.UIDB/50021/2020).

%
\bibliographystyle{iclr2023_conference}
\bibliography{references}

\nocite{brockman_openai_2016}

\appendix

\section{DQN algorithm pseudocode}

\begin{algorithm}
    \caption{DQN algorithm}\label{alg:dqn}
    \SetKwFunction{STEP}{env.step}
    \SetKwFunction{Reset}{env.reset}
    \SetKwFunction{Init}{Memory.init}
    \SetKwFunction{Add}{Memory.add}

    \For{$episode \leftarrow 1$ \KwTo $M$}{
        \For{$t\leftarrow 1$ \KwTo $T$}{
            \CommentSty{With probability $\epsilon$: $a_t=random()$, otherwise: $a_t = argmax_{a^\prime}\ Q(s,a^\prime)$}\;
            \CommentSty{Execute $a_t$ and observe $s_t^\prime$ and $r_t$}\;
            \CommentSty{Store transition $\{s_t,a_t, r_t, s_t^\prime\}$ in the replay buffer $\mathcal{D}$}\;
            \CommentSty{Sample a mini-batch of transitions $\{s_j,a_j,r_j,s^\prime_j\}$ from $\mathcal{D}$}\;
            \CommentSty{$y_j=r_j+\gamma\ max_{a^\prime_j}Q_\phi(s_j^\prime,a_j^\prime)$}\;
            \CommentSty{$\phi \leftarrow \phi - \alpha \sum_j\frac{dQ_\phi(s_j,a_j)}{d\phi}(Q_\phi(s_j,a_j)-y_j)$}\;
        }
    }
\end{algorithm}

\section{ViT patch size study}
\label{sec:vit_patch}

The patch size of a Vision Transformer can largely affect the performance of the model and the number
of computations per data sample. A patch size of 1 is equivalent to using the pixels as tokens while a
patch size of 16 converts patches of 16x16 pixels to a single token, i.e. the hyperparameter affects the
number of tokens quadratically. On the other hand, larger patches might not allow the model to learn
as good representations, therefore it is necessary to find a patch size that balances the computational
cost with task performance. For this work, we explored several different sizes and evaluated their data
efficiency by training Rainbow in the Atari game MsPacman for 100k across 10 different seeds. Our
results, Table \ref{tab:patch_size_eval}, show marginal differences between a patch size of 8 and 10, however, we didn't
observe a significant difference in the training time and for that reason, we decided to use a patch size of
8 for all our experiments.

\begin{table}[H]
	\centering
	\small
	\begin{tabular}{lll}
		\toprule
   		Patch size  & Score & Mean Time \\
		\midrule
		  6		        & 305.7 ± 71.0 &  4:56.33 \\
		  8		        & 801.9 ± 523.9 & 3:22.4 \\
		  10	        & 778.0 ± 324.0 & 3:10.2 \\
		  12	        & 627.0 ± 284.0 & 3:12.8 \\
		\bottomrule
	\end{tabular}
	\caption{Scores obtained by training Rainbow, using different path sizes for ViT, in MsPacman for 100k steps across 10 different seeds }
	\label{tab:patch_size_eval}
\end{table}

\begin{figure}[H]
	\centering
    \includegraphics[width=12cm]{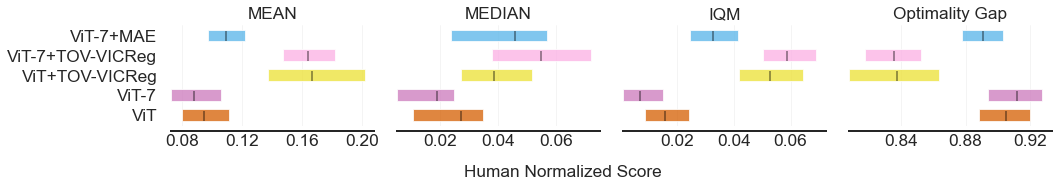}
	\caption{Comparision between a patch size of 7 and a patch size 8}
	\label{fig:seven_eight}
\end{figure}

However, MAE requires an image size, in our case 84x84, that is divisible by the patch size. The closest value to 8 is 7 which shows a very similar performance with and without pretraining, as shown in Figure \ref{fig:seven_eight}. However, using a patch size of 7 significantly increases the computation time of the experiments and for that reason, we decided to keep the patch size for the remaining methods.

\section{ResNet architecture}
\label{sec:resnet}

\begin{figure}[H]
	\centering
    \includegraphics[width=10cm]{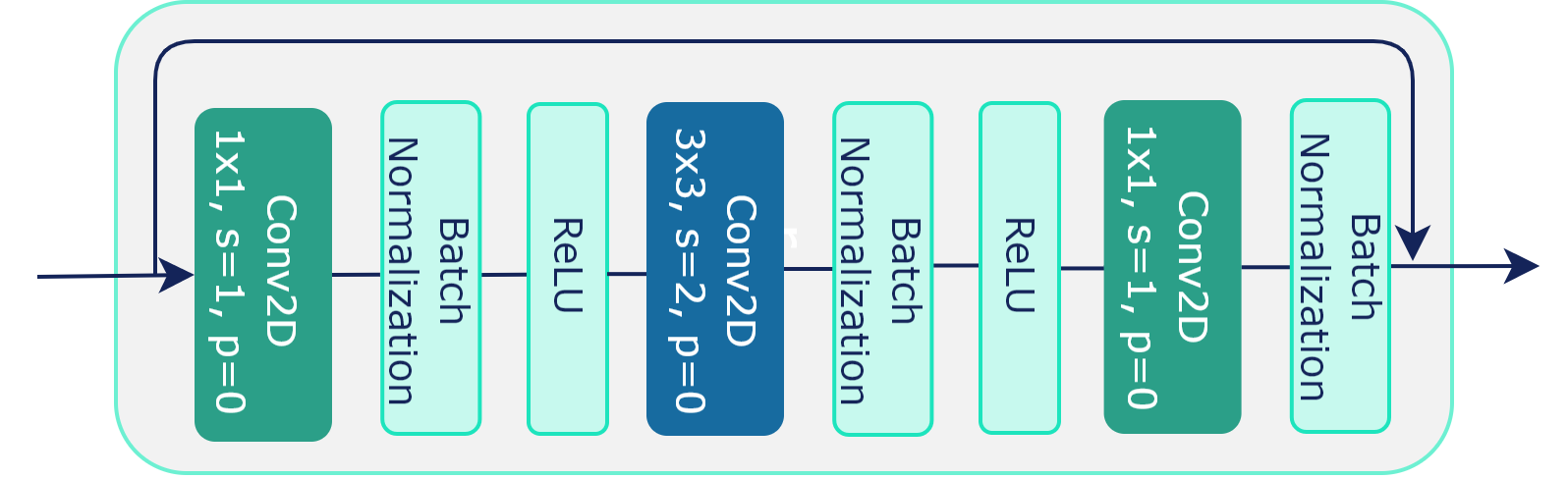}
	\caption{ResNet residual block}
	\label{fig:resnet}
\end{figure}

The ResNet we used is based on the ResNet used for SGI, which uses three inverted residual blocks with an expansion ratio of two, where each block is a sequence of Conv2D, Batch Normalization, and ReLU, as shown in Figure \ref{fig:resnet}.  However, to have a number of parameters similar to the ViT tiny we added an additional residual block and changed the channels of each block to 64, 128, 256 and 512. Additionally, we change the strides of each block to 2 for all blocks. The encoder computes representations vectors with a size of 18432.

\section{TOV-VICReg Pseudocode}
\label{sec:tov-vicreg-pseudo}
\begin{lstlisting}[language=Python, caption=Pytorch-like TOV-VICReg pseudocode]
# N: batch size, D: dimension of the embedding
# mse_loss: Mean square error loss function, off_diagonal: off-diagonal elements of a matrix, relu: ReLU activation function
# shuffle: shuffles elements in a certain dimension according to a permutation index
for u, v, w in loader: # load a batch with N samples
	# u -> x_{t}
	# v -> x_{t-1}
	# w -> x_{t+1}

	# apply augmentations
	u_a  = augmentation_1(u)
	u_b  = augmentation_2(u)
	v  = augmentation_3(v)
	w  = augmentation_3(w)

	# compute representations
	y_u_a = encoder(u_a)
	y_u_b = encoder(u_b)
	y_v = encoder(v)
	y_w = encoder(w)

	# compute embeddings
	z_u_a = expander(y_u_a)
	z_u_b = expander(y_u_b)
	z_v = expander(y_v)
	z_w = expander(y_w)

	shuffle_indexes = randint(0, 6) # sample from 0 to 3 permutations of 3
	labels = where(shuffle_indexes == 0, 0, 1)

	# concat and shuffle (N, 3, D)
	c = concat(p_u_a, p_v, p_w)
	c = shuffle(c, shuffle_indexes, dim=1)

	# temporal loss
	preds = linear(c) # Linear layer Dx6
	temp_loss = Binary_Cross_Entropy_Loss(preds, labels)

	# invariance loss
	sim_loss = mse_loss(z_a, z_b)

	# variance loss
	std_z_a = torch.sqrt(z_a.var(dim=0) + 1e-04)
	std_z_b = torch.sqrt(z_b.var(dim=0) + 1e-04)
	std_loss = torch.mean(relu(1 - std_z_a)) + torch.mean(relu(1 - std_z_b))

	# covariance loss
	z_a = z_a - z_a.mean(dim=0)
	z_b = z_b - z_b.mean(dim=0)
	cov_z_a = (z_a.T @ z_a) / (N - 1)
	cov_z_b = (z_b.T @ z_b) / (N - 1)
	cov_loss = off_diagonal(cov_z_a).pow_(2).sum() / D + \
			   off_diagonal(cov_z_b).pow_(2).sum() / D

	# loss
	loss = inv_coef * inv_loss + var_coef * var_loss + cov_coef * cov_loss + temp_coef * temp_loss

	# optimization step
	loss.backward()
	optimizer.step()
\end{lstlisting}

\section{TOV-VICReg augmentations}
\label{app:aug}
\begin{lstlisting}[language=Python, caption=Pytorch-like pseudocode of TOV-VICReg augmentations]
# Augmentation 1 / tau
RandomResizedCrop(84, scale=(0.08, 1.)),
RandomApply([
	ColorJitter(0.4, 0.4, 0.2, 0.1)
], p=0.8),
RandomGrayscale(p=0.2),
RandomApply([GaussianBlur((7, 7), sigma=(.1, .2))], p=1.0),
RandomHorizontalFlip()

# Augmentation 2 / tau prime
RandomResizedCrop(84, scale=(0.08, 1.)),
RandomApply([
	ColorJitter(0.4, 0.4, 0.2, 0.1)
], p=0.8),
RandomGrayscale(p=0.2),
RandomApply([GaussianBlur((7, 7), sigma=(.1, .2))], p=0.1),
RandomSolarize(120, p=0.2),
RandomHorizontalFlip(),

# Augmentation 3 / tau two prime and tau three prime
RandomApply([
	ColorJitter(0.4, 0.4, 0.2, 0.1)
], p=0.8),
RandomGrayscale(p=0.2),

\end{lstlisting}

\section{Rainbow implementation}
\label{section:der_comparison}

We trained our agents using a PyTorch implementation of the Rainbow algorithm available on GitHub, which offers enough flexibility to adapt it to our needs. In Table \ref{tab:der_comparison} we present a comparison between the implementation used and the official results reported by DER \citep{van_hasselt_when_2019}, we observed a similar performance in most games except for Assault, and Frostbite, where the official results are significantly higher. Despite these differences, we validated the implementation code and are confident that the results here presented are trustworthy. To allow the agents to play the Atari games we used the gym library \citep{brockman_openai_2016}, where for all games we used version number four of the environments (v4), disabled the default frame skip, and wrapped it with the DQN wrappers.

\begin{table}[h!]
	\centering
	\small
	\begin{tabular}{lll}
		\toprule
   		Game 		    & DER    & DER (ours) \\
		\midrule

		Alien 			& 739.9  & 446.6 ± 224.7 \\
		Assault         & 431.2  & 178.7 ± 87.1 \\
		Bank Heist 		& 51.0   & 23.8 ± 14.3 \\
		Breakout 		& 1.9    & 1.93 ± 1.43 \\
		Chopper Command & 861.8  & 696.0 ± 274.6 \\
		Freeway 		& 27.9   & 27.8 ± 2.0 \\
		Frostbite 		& 866.8  & 127.7 ± 25.8 \\
		Kangaroo 		& 779.3  & 448.0 ± 648.0 \\
		MsPacman 		& 1204.1 & 1015 ± 487.1 \\
		Pong 			& -19.3  & -18.6 ± 4.4 \\

		\bottomrule
	\end{tabular}
	\caption{Comparison between DER scores and our implementation scores}
	\label{tab:der_comparison}
\end{table}

\section{Atari Environments setup}

We used the Atari games available at the gym library \citep{brockman_openai_2016} (version 0.23.1), and all games were run using their 4th version without frame skip, e.g. "AlienNoFrameskip-v4". Furthermore, we employ similar wrappers to the environments as previous works \citep{mnih_human-level_2015}, namely, scaled observation to 84x84, changed observations to grayscale, stacked observations, applied a max number of no-op actions, and terminated the environment when the agent loses a life.

\begin{lstlisting}[language=Python, caption=Gym Atari Wrappers]
	env = AtariPreprocessing(env, terminal_on_life_loss=True, scale_obs=True)
    env = TransformReward(env, np.sign)
    env = FrameStack(env, 3)
\end{lstlisting}

\section{Self-Supervised methods hyperparameters}
\label{section:hyper}

\begin{table}[H]
	\centering
	\small
	\begin{tabular}{lll}
		\toprule
   		Hyperparameter 		    & Value \\
		\midrule

		Drop path rate 			& 0.1 \\
		Freeze last layer       & True \\
		\# local crops 		    & 8 \\
		Local crops scale interval & [0.05, 0.5] \\
		Learning rate 			& $5.0 \times 10^{-4}$ \\
		Min learning rate 		& $1.0 \times 10^{-6}$ \\
		Teacher ema coefficient & 0.996 \\
		Normalize last layer 	& False \\
		Optimizer 				& AdamW \\
		Out dimension 			& 1024 \\
		Use batch normalization in head & false \\
		Teacher warmup temperature & 0.04 \\
		\# warmup epochs for teacher temperature & 0 \\
		Weight decay 			& 0.04 \\
		Weight decay final value & 0.4 \\

		\bottomrule
	\end{tabular}
	\caption{DINO hyperparameters}
	\label{tab:dino_hyper}
\end{table}

\begin{table}[H]
	\centering
	\small
	\begin{tabular}{lll}
		\toprule
   		Hyperparameter 		    & Value \\
		\midrule

		Random crop min scale 	& 0.08 \\
		Learning rate			& 0.6 \\
		Number of features 		& 256 \\
		Momentum encoder ema coefficient & 0.99 \\
		MLP hidden dimensions 	& 4096 \\
		Softmax temperature		& 1.0 \\
		Optimizer				& LARS \\
		Weight decay			& $1.0 \times 10^{-6}$ \\
		\bottomrule
	\end{tabular}
	\caption{MoCo v3 hyperparameters}
	\label{tab:moco_hyper}
\end{table}

\begin{table}[H]
	\centering
	\small
	\begin{tabular}{lll}
		\toprule
   		Hyperparameter 		    & Value \\
		\midrule

		Base Learning Rate		& 0.2 \\
		Covariance coefficient	& 1.0 \\
		MLP dimensions			& 1024-1024-1024 \\
		Invariance coefficient	& 25.0 \\
		Variance coefficient 	& 25.0 \\
		Weight decay			& $1.0 \times 10^{-6}$ \\
		\bottomrule
	\end{tabular}
	\caption{VICReg hyperparameters}
	\label{tab:vicreg_hyper}
\end{table}

\begin{table}[H]
	\centering
	\small
	\begin{tabular}{lll}
		\toprule
   		Hyperparameter 		    & Value \\
		\midrule

		Base Learning Rate		& 0.6 \\
		Covariance coefficient	& 10.0 \\
		MLP dimensions			& 1024-1024-1024 \\
		Invariance coefficient	& 25.0 \\
		Variance coefficient 	& 25.0 \\
		Weight decay			& $1.0 \times 10^{-6}$ \\
		\bottomrule
	\end{tabular}
	\caption{TOV-VICReg hyperparameters}
	\label{tab:tov_vicreg_hyper}
\end{table}

\section{Models used}

\begin{table}[H]
	\centering
	\begin{tabular}{ll}
		\toprule
		Model Name  & \# parameters \\
		\midrule
		Nature CNN  & 75.936 \\
		ResNet & 4.932.524 \\
		ViT tiny & 5.526.720 \\
		\bottomrule
	\end{tabular}
	\caption{Number of learnable parameters of each model we used}
	\label{tab:model_size}
\end{table}

\section{Results Table}

\begin{table}[H]
	\centering
	\tiny
	\begin{tabular}{p{0.1\textwidth}p{0.07\textwidth}p{0.07\textwidth}p{0.07\textwidth}p{0.07\textwidth}p{0.07\textwidth}p{0.07\textwidth}p{0.07\textwidth}p{0.07\textwidth}}
		\toprule
				Games 	&     Nature CNN &    ResNet &           ViT & ViT+TOV-VICReg &      ViT+DINO &      ViT+MoCo &    ViT+VICReg & ViT-7+MAE \\
		\midrule
			Assault 	&  355.1 ± 105.2 & 452.0 ± 349.1 & 322.7 ± 146.9 &  366.3 ± 124.5 & \textbf{493.3 ± 254.7} & 493.3 ± 181.1 & 408.5 ± 156.6		& 507.4 ± 381.4 \\
			Alien		&  210.8 ± 133.1 & 186.9 ± 104.4 & 250.6 ± 142.6 &  197.6 ± 114.4 & 275.1 ± 153.2 & \textbf{380.8 ± 194.7} & 187.3 ± 118.8 		& 263.13 ± 158.7\\
			Bank Heist	&    37.6 ± 29.5 &   30.6 ± 18.6 &  \textbf{58.3 ± 115.4} &    34.5 ± 18.8 &   18.6 ± 10.7 &   21.0 ± 30.1 &   29.6 ± 13.6 		& 13.1 ± 6.4	\\
			Breakout 	&      \textbf{5.1 ± 3.3} &     4.7 ± 2.1 &     3.2 ± 2.6 &      4.3 ± 2.7 &     2.8 ± 2.1 &     2.7 ± 1.6 &     3.1 ± 1.6		& 1.88 ± 1.6	\\
			Chopper Command &  828.0 ± 323.8 & 737.0 ± 354.0 & 747.0 ± 268.5 &  \textbf{853.0 ± 312.2} & 760.0 ± 249.0 & 968.0 ± 673.0 & 668.0 ± 274.9 	& 809.0 ± 284.6	\\
			Freeway 	&     \textbf{30.4 ± 1.2} &    26.5 ± 2.5 &    21.2 ± 1.4 &     25.9 ± 2.7 &    25.0 ± 2.0 &    22.5 ± 2.1 &    23.7 ± 2.4 		& 22.8 ± 2.0 	\\
			Frostbite 	&   120.1 ± 25.9 &  107.9 ± 26.8 &  127.5 ± 15.6 &  \textbf{143.7 ± 106.7} &  132.7 ± 14.1 &  111.3 ± 37.0 &  120.0 ± 18.2 		& 157.3 ± 216.6 \\
			Kangaroo 	& \textbf{776.0 ± 1035.4} & 405.0 ± 226.4 &   60.0 ± 91.7 & 704.0 ± 1076.7 & 316.0 ± 233.5 & 384.0 ± 531.0 & 268.0 ± 244.5 		& 229.0 ± 66.8  \\
			MsPacman 	&  \textbf{781.3 ± 417.1} & 757.7 ± 413.2 & 618.9 ± 259.9 &  639.5 ± 378.4 & 698.9 ± 374.5 & 586.4 ± 257.5 & 633.0 ± 372.1		& 753.8 ± 531.6 \\
				Pong 	&    -13.6 ± 9.7 &   -12.0 ± 8.6 &   -21.0 ± 0.0 &    \textbf{-6.2 ± 13.4} &   -18.4 ± 3.4 &   -17.6 ± 4.3 &   -15.1 ± 3.9 		& -18.9 ± 1.7	\\
		\bottomrule
	\end{tabular}
	\caption{Table of results (mean and standard error) from experiments presented in Section \ref{section:data-efficiency}.The bold values represent the best scores for the corresponding game}
	\label{tab:results}
\end{table}

\section{Data-efficiency in unseen environments}

Table \ref{tab:unseen} shows a comparison of the randomly initialized and a pre-trained (using TOV-VICReg) Vision Transformer in Atari games that were not used in the pre-training phase. In general, both models seem to perform very similarly as indicated by the IQM over the aggregated normalized scores. However, in RoadRunner the pretraining seems to degrade data-efficiency and in Venture the pretraining seems to improve data-efficiency. In short, we don't find any advantage in using a pre-trained vision transformer for games that were not used during pretraining. We don't find this result surprising given the lack of variety present in the dataset used for pretraining which reduces the possibility of the encoder finding features that can be used elsewhere.\\

\begin{table}[H]
    \centering
\begin{tabular}{lll}
	\toprule
			Games &        ViT & TOV-VICReg+ViT \\
	\midrule
		  Asterix &   443.5 ± 225.6 &  445.0 ± 214.9 \\
			Krull &   944.5 ± 525.8 &  708.9 ± 572.3 \\
	   RoadRunner & 2687.0 ± 2884.3 & 913.0 ± 1289.9 \\
	SpaceInvaders &   184.3 ± 117.0 &   155.9 ± 91.0 \\
		  Venture &      4.0 ± 28.0 &   76.0 ± 152.4 \\
    \midrule
        IQM       &     0.0174      & 0.0186 \\
	\bottomrule
\end{tabular}
	\caption{Mean and standard error results of the evaluations across 10 different training runs, where at each evaluation the agent plays 10 episodes of the game. The agent was trained using the Rainbow algorithm for 100k steps.}
	\label{tab:unseen}
\end{table}

\section{Linear Probing Evaluation Task}
\label{section:eval_complete}

Evaluating representations computed by a pretrained encoder is a difficult task. One possible option is assessing improvements in data efficiency in a reinforcement learning task, as we did in the previous section. However, the results usually suffer from a high level of uncertainty which requires us to run dozens of training runs, thus making it computationally expensive. Another possible path would be using previously proposed benchmarks like the AtariARI benchmark \cite{anand_unsupervised_2020}, which tries to evaluate representations using the RAM states as ground truth labels. However, this only works for 22 Atari games (out of 62) and requires the encoder to use the full observation provided by the environments (160x210). For those reasons, we use a different evaluation task that is more efficient, allowing us to test more pretrained models during the research process ($\sim 50$ min per game), and flexible, meaning that we can use it in different environments.
Our evaluation task is a simple Imitation Learning task where we train a network, composed of a frozen pre-trained encoder and a linear layer, i.e. linear probing, to correctly predict the action that a certain policy will perform given its current observation. The intuition to use such an evaluation is that a representation that allows an agent to efficiently learn an environment must encode state information that can be recovered by a linear layer and which can be used to learn other tasks efficiently.

We present the results in Table \ref{tab:eval_complete}, we compare against a random classifier, i.e. uniform sampling, randomly initialized networks and a non-frozen encoder which we use as a goal score. All methods were trained for 100 epochs except the latter which we trained for 300. We use the DQN Replay dataset to obtain the observations and the actions we obtain the datapoints from the last checkpoint of each game, where we consider the policy to be less stochastic. The train dataset is composed of 100 thousand observations from the game we are testing and the test dataset is composed of 10 thousand. ViT+TOV-VICReg L corresponds to a ViT tiny pretrained with TOV-VICReg on the 26 Atari games from the Atari100k.

\begin{table}
    \centering
    \tiny
    \begin{tabular}{p{0.07\textwidth}p{0.059\textwidth}|p{0.059\textwidth}p{0.059\textwidth}p{0.059\textwidth}|p{0.059\textwidth}p{0.059\textwidth}p{0.059\textwidth}p{0.059\textwidth}p{0.059\textwidth}|p{0.059\textwidth}}
        \toprule
        & & \multicolumn{3}{|c|}{Randomly initialized encoder}  & \multicolumn{5}{c|}{Pre-trained encoder} & \multicolumn{1}{c}{W/o freeze} \\
        \midrule
		& Random Classifier & Nature CNN & ResNet &    ViT & ViT+TOV-VICReg & ViT+DINO & ViT+MoCo & ViT+VICReg & ViT+TOV-VICReg L & Nature CNN \\
		Game           &                   &            &        &        &                &          &          &            &                  &           \\
		\midrule
		Alien          &            0.0556 &     0.0077 & 0.0558 & 0.0147 &         0.1003 &   0.0470 &   0.0646 &     0.0695 &           0.0988 & \textbf{0.1021} \\
		Assault        &            0.1519 &     0.1497 & 0.2270 & 0.1770 &         0.3044 &   0.2536 &   0.2557 &     0.3704 &           0.3065 & \textbf{0.6673} \\
		BankHeist      &            0.0608 &     0.0780 & 0.1312 & 0.0756 &         0.1622 &   0.1059 &   0.1083 &     0.1467 &           0.1523 & \textbf{0.2080} \\
		Breakout       &            0.2509 &     0.1311 & 0.3850 & 0.2183 &         0.3285 &   0.3591 &   0.2765 &     0.4077 &           0.3099 & \textbf{0.5907} \\
		Chopper Command &            0.0563 &     0.0145 & 0.0647 & 0.0176 &         \textbf{0.3225} &   0.0383 &   0.2019 &     0.1298 &           0.3088 & 0.2660 \\
		Freeway        &            0.3999 &     0.6808 & 0.6850 & 0.6843 &         0.7041 &   0.6850 &   0.6972 &     0.6971 &           0.6942 & \textbf{0.8885} \\
		Frostbite      &            0.0565 &     0.0302 & 0.0730 & 0.0367 &         \textbf{0.1021} &   0.0517 &   0.0744 &     0.0664 &           0.1001 & 0.1019 \\
		Kangaroo       &            0.0603 &     0.0311 & 0.1039 & 0.0562 &         0.2184 &   0.0877 &   0.1374 &     0.1259 &           0.2126 & \textbf{0.3311} \\
		MsPacman       &            0.1121 &     0.0388 & 0.1419 & 0.0780 &         0.1527 &   0.1215 &   0.1168 &     0.1400 &           0.1500 & \textbf{0.2063} \\
		Pong           &            0.1644 &     0.0692 & 0.1702 & 0.0718 &         0.2853 &   0.1447 &   0.2730 &     0.2337 &           0.3042 & \textbf{0.4340} \\
		\midrule
		Mean           &            0.1369 &     0.1231 & 0.2038 & 0.1430 &         0.2680 &   0.1894 &   0.2206 &     0.2387 &           0.2637 & \textbf{0.3796} \\
		\bottomrule
    \end{tabular}
    \caption[Evaluation task results]{F1-scores for each game evaluated and mean. We trained all the encoders in all games seperatly for 100 epochs over a dataset of 100k observations and evaluate in 10k new observations. The rightmost column show the results of a Nature CNN encoder that was not frozen during train and which we use as a goal for the remaining.}
    \label{tab:eval_complete}
\end{table}

\begin{figure}[h]
	\centering
    \includegraphics[width=8cm]{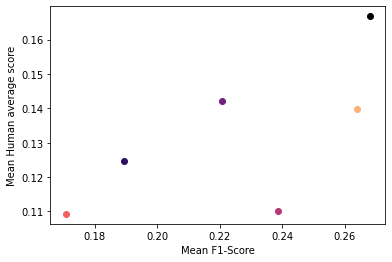}
	\caption[RL and Evaluation task relation]{Relation between the mean average human score obtained in RL and the mean F1-score obtained in the evaluation task of several experiments}
	\label{fig:eval-corr}
\end{figure}

To validate our evaluation task we calculate the Pearson correlation coefficient between the mean of the average human normalized scores, obtained in the reinforcement learning, and the mean of the F1-scores, from the evaluation task of all pretrained models. We report a Pearson correlation factor of 0.6794. Even though we are not in the presence of a strong correlation there is a clear trend for the RL scores to increase when the evaluation scores also increase, as observed in Figure \ref{fig:eval-corr}. Despite the promising results, more data points are needed, especially using different pre-training methods, which would allow us to better validate this evaluation task. Nevertheless, we believe that the evaluation task might be a compelling tool for future methods that try to learn good representations for a reinforcement learning task.

\end{document}